\documentclass[11pt,a4paper]{article}
\pdfoutput=1

\usepackage[utf8]{inputenc}
\usepackage[T1]{fontenc}
\usepackage{lmodern}
\usepackage[english]{babel}
\usepackage[a4paper,top=1in,bottom=1in,left=1.1in,right=1.1in]{geometry}
\usepackage{amsmath,amssymb,amsthm}
\usepackage{booktabs}
\usepackage{siunitx}
\usepackage{url}
\usepackage{graphicx}
\usepackage{subcaption}
\usepackage{listings}
\usepackage{xcolor}
\usepackage{authblk}
\usepackage{float}
\usepackage{titlesec}
\usepackage{parskip}
\usepackage[hidelinks,colorlinks=true,linkcolor=blue!40!black,citecolor=blue!50!black,urlcolor=blue!60!black]{hyperref}
\usepackage[capitalise,nameinlink]{cleveref}
\usepackage{microtype}

\titleformat{\section}{\normalfont\Large\bfseries\sffamily}{\thesection}{0.7em}{}
\titleformat{\subsection}{\normalfont\large\bfseries\sffamily}{\thesubsection}{0.6em}{}
\titleformat{\subsubsection}{\normalfont\normalsize\bfseries\sffamily}{\thesubsubsection}{0.5em}{}
\titleformat{\paragraph}[runin]{\normalfont\normalsize\bfseries\sffamily}{}{0pt}{}[\hskip .5em]
\titlespacing*{\paragraph}{0pt}{0.6\baselineskip}{0.5em}
\titlespacing*{\section}{0pt}{1.2\baselineskip}{0.5\baselineskip}
\titlespacing*{\subsection}{0pt}{1.0\baselineskip}{0.4\baselineskip}

\definecolor{codebg}{RGB}{248,248,248}
\definecolor{codekw}{RGB}{0,76,153}
\definecolor{codestr}{RGB}{163,21,21}
\definecolor{codecom}{RGB}{0,110,0}

\lstdefinestyle{pycode}{
  language=Python,
  basicstyle=\ttfamily\footnotesize,
  keywordstyle=\color{codekw}\bfseries,
  commentstyle=\color{codecom}\itshape,
  stringstyle=\color{codestr},
  showstringspaces=false,
  numbers=left,
  numberstyle=\tiny\color{gray},
  stepnumber=1,
  numbersep=6pt,
  frame=single,
  framerule=0.4pt,
  rulecolor=\color{black!20},
  backgroundcolor=\color{codebg},
  breaklines=true,
  breakatwhitespace=true,
  postbreak=\mbox{\textcolor{gray}{$\hookrightarrow$}\space},
  tabsize=4,
  keepspaces=true,
  columns=fullflexible,
  aboveskip=10pt,
  belowskip=10pt,
  xleftmargin=12pt,
  xrightmargin=0pt
}

\newcommand{\pycsp}{\texorpdfstring{PyCSP$^3$}{PyCSP3}}
\newcommand{\xcsp}{\texorpdfstring{XCSP$^3$}{XCSP3}}
\newcommand{\pycspsched}{\texorpdfstring{\pycsp{}-Scheduling}{PyCSP3-Scheduling}}

\title{\pycspsched{}: A Scheduling Extension for \pycsp{}}
\author[1]{Sohaib Afifi\thanks{\href{mailto:sohaib.afifi@univ-artois.fr}{\texttt{sohaib.afifi@univ-artois.fr}}}}
\affil[1]{Univ. Artois, UR 3926, Laboratoire de G\'enie Informatique et d'Automatique de l'Artois (LGI2A), F-62400 B\'ethune, France}
\date{\today}

\hypersetup{
  pdftitle  = {PyCSP3-Scheduling: A Scheduling Extension for PyCSP3},
  pdfauthor = {Sohaib Afifi},
  pdfkeywords = {Constraint Programming, Modelling Languages, Scheduling, PyCSP3, XCSP3}
}

\begin{document}
\maketitle

\begin{abstract}
\pycsp{} provides a productive way to build constraint models for solving combinatorial constrained problems and export them to \xcsp{}, preserving a complete separation between modeling and solving.
However, it lacks native support for scheduling abstractions such as interval variables, sequence variables, and resource functions.
As a result, scheduling models must be encoded with low-level integer variables and manual channeling constraints, even though \pycsp{} already provides global constraints like \texttt{NoOverlap} and \texttt{Cumulative} on integer arrays.
We present \pycspsched{}, a library that adds scheduling abstractions to \pycsp{} through 53 dedicated constraints and 27 expressions, and compiles them down to standard \pycsp{}/\xcsp{} constraints, maintaining the modeling/solving separation that underpins the \pycsp{} ecosystem.
On 261 paired instances across 17 model families (5 runs each), both formulations produce identical objectives on all 72 doubly-proved optimal pairs and nearly half of the families (8/17) remain structurally unchanged after compilation; however, runtime performance diverges across families, with clear gains on some (up to 5.8\(\times\)) and regressions on others due to the overhead of compilation decompositions.

\noindent\textbf{Source code and benchmarks:} \url{https://github.com/sohaibafifi/pycsp3-scheduling}
\end{abstract}

\section{Introduction}
Consider a typical job-shop model in \pycsp{}~\cite{pycsp3_2020}: the modeler declares integer start-time variables, writes arithmetic precedence constraints (\(s_i + d_i \le s_j\)), and posts pairwise disjunctions for machine conflicts.
\pycsp{} already provides global constraints \texttt{NoOverlap} and \texttt{Cumulative} that operate on integer arrays~\cite{pycsp3_2020}, but the modeler must still manage start-time variables, duration arrays, and resource-height lists explicitly, with no notion of optionality or sequencing.
Adding optional tasks requires separate Boolean variables and implication guards on every temporal constraint; setup times require yet another layer of conditional arithmetic.
These encodings are correct, but they obscure the scheduling structure and must be rebuilt from scratch for each new model.

Industrial CP systems such as CP Optimizer~\cite{laborie2009cpoptimizer}, MiniZinc~\cite{minizinc2007}, and OR-Tools~\cite{ortools} address this by providing scheduling objects, but they either couple models to a single (often commercial) solver or lack first-class scheduling variable types (Section~\ref{sec:related}).

\pycspsched{} fills this gap for the \pycsp{}/\xcsp{} ecosystem.
It provides \texttt{IntervalVar}, \texttt{SequenceVar}, precedence/overlap constraints, and resource functions, all compiled down to standard \pycsp{} variables and constraints (including existing globals like \texttt{NoOverlap} and \texttt{Cumulative} when applicable) and exported as \xcsp{} output.
The library is open source (MIT licence), preserves the complete separation between modeling and solving~\cite{pycsp3_2020}, and produces standardized \xcsp{}~\cite{xcsp3_core_2020,xcsp3_integrated_2020} instances solvable by any compatible solver.

This paper makes 3 contributions:
\begin{enumerate}
\item A scheduling API for \pycsp{} centered on interval and sequence abstractions, with support for optionality, intensity functions, and setup-time-aware sequencing.
\item A compilation scheme that lowers these abstractions to \pycsp{}/\xcsp{} constraints while preserving satisfiability and optimality.
\item An empirical comparison of 261 paired instances (classical \pycsp{} formulation vs.\ scheduling formulation), showing semantic consistency on solved optimal pairs and characterizing the runtime/size tradeoffs across 17 model families.
\end{enumerate}

\section{Related Work}
\label{sec:related}
Scheduling abstractions in CP can be organized along 2 design axes: whether the abstraction is \emph{solver-integrated} or \emph{compilation-based}, and whether the modeling language is \emph{standalone} or \emph{embedded} in a host language.

CP Optimizer~\cite{ibmcpoptimizer,laborie2009cpoptimizer} and its Python API DOcplex.CP~\cite{docplexcp} exemplify the solver-integrated approach: interval variables are first-class solver objects with dedicated propagation.
This yields strong propagation, including for optional intervals and transition-aware disjunctive resources, but couples models to a single (commercial) solver.
MiniZinc~\cite{minizinc2007,mznchallenge2023} takes the compilation-based, standalone-language approach: it provides scheduling globals (\texttt{cumulative}, \texttt{disjunctive}, \texttt{span}), but no first-class interval or sequence variable types; the modeler must decompose scheduling structures into individual integer start-time and duration variables.
A MiniZinc record such as \texttt{\{start: var int, duration: int\}} can group such fields, but it remains a passive aggregation of integers: it does not by itself carry optional-interval semantics, automatically guard temporal/resource constraints by presence, expose sequence-level accessors such as \texttt{next}/\texttt{previous}, or trigger the emission of scheduling globals over decompositions.
OR-Tools CP-SAT~\cite{ortools} provides an embedded Python API with scheduling helpers (interval and disjunctive variables, cumulative resources), but these are tied to the CP-SAT solver and not exchangeable with other CP back-ends.

\pycspsched{} occupies a different cell in this design space: it is \emph{compilation-based}, \emph{embedded} in \pycsp{}, and targets an open exchange format (\xcsp{}).
To our knowledge, it is the first scheduling layer that provides first-class \texttt{IntervalVar}/\texttt{SequenceVar} abstractions with optionality and transition-aware sequencing while compiling to a solver-independent standard rather than a single back-end.
\pycsp{}~\cite{pycsp3_2020} already provides \texttt{NoOverlap} and \texttt{Cumulative} as global constraints on integer-variable arrays, and uses them in scheduling examples (Flow Shop, RCPSP); however, these remain low-level: the modeler manages start-time arrays, duration lists, and resource-height vectors manually, with no support for optionality, intensity, or sequencing.
\pycspsched{} lifts these into higher-level abstractions and compiles back down to \pycsp{} constraints (reusing the existing globals when possible), producing \xcsp{}~\cite{xcsp3_core_2020,xcsp3_integrated_2020} output that can be solved by a broad ecosystem of solvers: ACE~\cite{ace_solver}, Choco~\cite{choco_solver}, CoSoCo~\cite{cosoco_solver}, OR-Tools~\cite{ortools}, CP Optimizer~\cite{ibmcpoptimizer}, Z3~\cite{z3_solver}, Pumpkin~\cite{pumpkin_solver}, as well as MiniZinc backends via conversion~\cite{pycsp3solversextra}.
This solver diversity is a key advantage: existing \pycsp{} model code, tooling, and solver bindings remain usable without modification.

The foundational CP scheduling literature~\cite{baptiste2001constraint,pinedo2016scheduling,rossi2006handbook} informs the API design, particularly the semantics of optional intervals, intensity functions, and transition-aware sequencing.

\section{System and Compilation}
\label{sec:system}

The library introduces 2 variable types (\texttt{IntervalVar}, \texttt{SequenceVar}), temporal constraints (precedence, no-overlap, grouping, forbidden-time), resource functions (cumulative, state), and accessor expressions (\texttt{start\_of}, \texttt{end\_of}, \texttt{presence\_of}).
The code below illustrates the difference on a simplified RCPSP fragment (compare with the \pycsp{} RCPSP model in~\cite{pycsp3_2020}).

\textbf{Classical formulation (\pycsp{}).}\par\smallskip
\noindent\begin{minipage}{\linewidth}
\begin{lstlisting}[style=pycode]
s = VarArray(size=nJobs, dom=range(horizon))

satisfy(
    [s[i] + durations[i] <= s[j]
        for i in range(nJobs) for j in successors[i]],
    [Cumulative(
        Task(origin=s[i], length=durations[i],
             height=quantities[i][k])
        for i in range(nJobs) if quantities[i][k] > 0
     ) <= cap for k, cap in enumerate(capacities)]
)

minimize(s[-1])
\end{lstlisting}
\end{minipage}

\textbf{Scheduling formulation (\pycspsched{}).}\par\smallskip
\noindent\begin{minipage}{\linewidth}
\begin{lstlisting}[style=pycode]
x = [IntervalVar(start=(0, horizon), size=durations[i])
     for i in range(nJobs)]

satisfy(
    [end_before_start(x[i], x[j])
        for i in range(nJobs)
        for j in successors[i]],
    [sum(pulse(x[i], quantities[i][k])
         for i in range(nJobs)
         if quantities[i][k] > 0) <= capacities[k]
     for k in range(nResources)]
    # alt: use SeqCumulative(intervals, heights, capacity)
)

minimize(end_of(x[-1]))
\end{lstlisting}
\end{minipage}

The scheduling formulation replaces manual \texttt{Task(origin=..., length=..., height=...)} constructions with \texttt{pulse(interval, height)}, and explicit arithmetic precedence with \texttt{end\_before\_start}.
An even higher-level alternative is available via \texttt{SeqCumulative(intervals, heights, capacity)}, which encapsulates the cumulative constraint in a single call.
Adding optionality or setup times changes 1 keyword argument, not the entire constraint structure.
This is more visible on harder cases.
For instance, in flexible job-shop the scheduling formulation drops from 120 to 83 lines (a 31\% reduction), and the contrast becomes more pronounced once optional alternative modes and machine-dependent setup times are added:

\noindent\begin{minipage}{\linewidth}
\begin{lstlisting}[style=pycode]
# Each operation may run on any of M alternative machines (mode = optional interval).
modes = [[IntervalVar(start=(0, H), size=durations[i][m], optional=True)
          for m in range(nMachines)]
         for i in range(nOps)]
op    = [IntervalVar(start=(0, H), size=(1, H)) for _ in range(nOps)]

# Each machine is a sequence of typed intervals with setup times.
machine = [SequenceVar([modes[i][m] for i in range(nOps) if durations[i][m] >= 0],
                       types=[op_type[i] for i in range(nOps) if durations[i][m] >= 0])
           for m in range(nMachines)]

satisfy(
    [alternative(op[i], modes[i]) for i in range(nOps)],         # exactly one mode runs
    [SeqNoOverlap(machine[m], transition_matrix=setup[m])         # type-dependent setups
        for m in range(nMachines)],
    [end_before_start(op[i], op[j]) for (i, j) in precedences],
)

minimize(makespan(op))
\end{lstlisting}
\end{minipage}

The classical encoding of the same model needs explicit Boolean mode-selection variables, manual presence-guarded precedences, manual disjunctive disjunctions per machine pair, and a hand-written setup-time term in every disjunct.
\texttt{optional=True}, \texttt{alternative}, and \texttt{transition\_matrix} fold these patterns into the abstractions.

\begin{figure}[t]
\centering
\begin{subfigure}[t]{\textwidth}
\centering
\includegraphics[width=\textwidth]{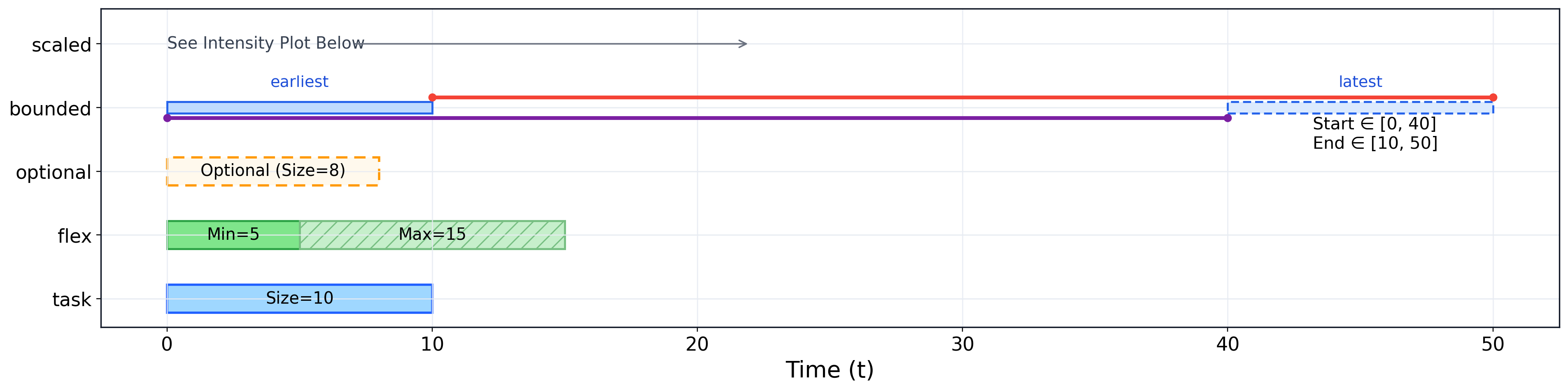}
\caption{Definitions of fixed, optional, flexible, bounded, and scaled interval variants.}
\end{subfigure}
\vspace{3pt}
\begin{subfigure}[t]{\textwidth}
\centering
\includegraphics[width=\textwidth]{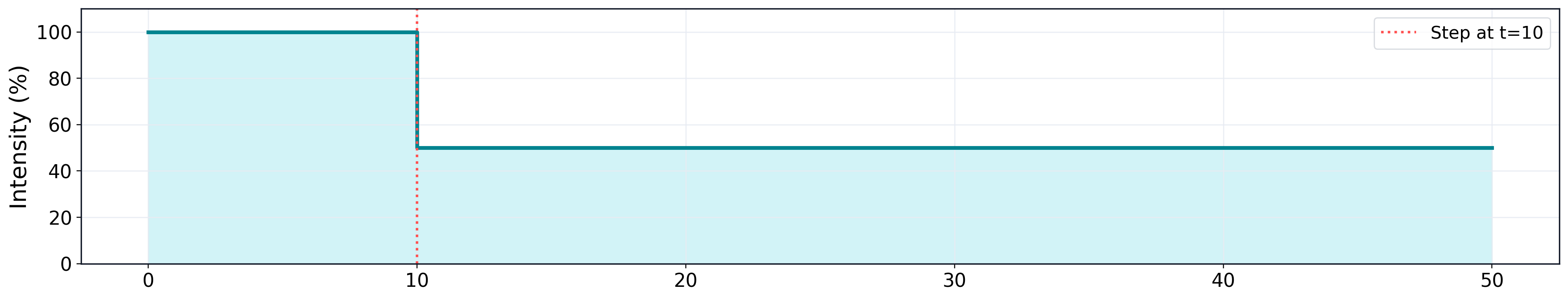}
\caption{Intensity profile used by the scaled interval example.}
\end{subfigure}
\vspace{3pt}
\begin{subfigure}[t]{\textwidth}
\centering
\includegraphics[width=\textwidth]{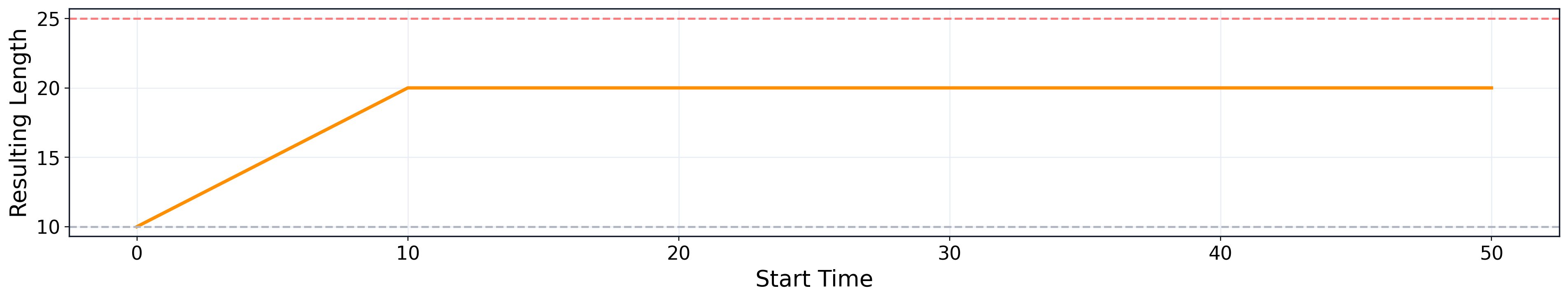}
\caption{Resulting duration as a function of start time under the intensity profile.}
\end{subfigure}
\caption{Interval-variable anatomy: definitions (top), intensity profile (middle), and resulting duration vs.\ start time (bottom).}
\label{fig:interval-docstyle}
\end{figure}

\paragraph{Interval variables.}
An \texttt{IntervalVar} represents a task with attributes start, end, size, length, and presence.
Each attribute may be a fixed integer or a domain (\texttt{size=5} or \texttt{size=(lb,ub)}; an existing \pycsp{} integer variable is also accepted), so duration can be a parameter, a bounded decision, or any pre-existing decision variable.
The library exposes 5 interval variants illustrated in Figure~\ref{fig:interval-docstyle} and accessible via the same constructor:
\begin{description}
\item[Fixed.] \texttt{IntervalVar(start=(0, H), size=5)} declares a task of length 5 placed within horizon \(H\); the start is decided by the solver.
\item[Flexible.] \texttt{IntervalVar(start=(0, H), size=(3, 7))} lets the solver pick both start and size within bounds; \texttt{length=(lb,ub)} is the analogous form when length and size differ (intensity case below).
\item[Optional.] \texttt{IntervalVar(start=(0, H), size=5, optional=True)} attaches a Boolean presence variable; all temporal and resource constraints involving the interval are automatically guarded by presence at compile time.
\item[Bounded.] Setting \texttt{start=(s\_lb, s\_ub)} and \texttt{end=(e\_lb, e\_ub)} jointly restricts release date and deadline; the constructor narrows the implied size domain accordingly.
\item[Scaled.] \texttt{IntervalVar(start=(0, H), size=10, intensity=profile, granularity=100)} couples the elapsed length to a stepwise efficiency \(\texttt{profile}(t)\), so a task of fixed work content needs more elapsed time when scheduled in a low-intensity region. Formally,
\begin{equation}
\texttt{size} \cdot \texttt{granularity} = \sum_{t=\texttt{start}}^{\texttt{start}+\texttt{length}-1} \texttt{intensity}(t).
\end{equation}
\end{description}
The same accessors (\texttt{start\_of}, \texttt{end\_of}, \texttt{size\_of}, \texttt{length\_of}, \texttt{presence\_of}) apply uniformly to all 5 variants and return \pycsp{} expressions that integrate with raw \pycsp{} constraints (Section~\ref{sec:escape-hatch}).

\paragraph{Sequence variables and arrays.}
A \texttt{SequenceVar} groups a list of interval variables into an ordered sequence, typically representing a disjunctive resource (machine).
Each interval in the sequence can be assigned an integer \emph{type} (an \emph{ID}), enabling type-dependent transition times in \texttt{SeqNoOverlap}.
Factory functions \texttt{IntervalVarArray} and \texttt{SequenceVarArray} create multi-dimensional collections, mirroring the array idioms of \pycsp{}.
These arrays accept expressions as indices (e.g., \texttt{array[next\_arg(seq, x)]}), enabling data access indexed by decision variables (see \textbf{element expressions}).

\paragraph{Constraints and expressions.}
Tables~\ref{tab:constraints} and~\ref{tab:expressions} list all constraints and expressions provided by the library.
The interval accessors \texttt{start\_of}, \texttt{end\_of}, \texttt{size\_of}, \texttt{length\_of}, and \texttt{presence\_of} return \texttt{IntervalExpr} objects that overload Python's comparison and arithmetic operators (\texttt{<}, \texttt{<=}, \texttt{>=}, \texttt{>}, \texttt{==}, \texttt{!=}, \texttt{+}, \texttt{-}, \texttt{*}, \texttt{//}, \texttt{abs}).
A precedence such as \texttt{end\_before\_start(a, b, delay)} can therefore equivalently be written as \texttt{start\_of(b) >= end\_of(a) + delay}.
The named precedence functions in the table are syntactic sugar over this operator form for the 8 combinations of \(\{\textit{at}, \textit{before}\} \times \{\textit{start}, \textit{end}\}^2\), kept for readability when modeling pure precedence networks.

\begin{table}[t]
\centering
\caption{Constraints provided by \pycspsched{}.}
\label{tab:constraints}
\footnotesize
\setlength{\tabcolsep}{4pt}
\begin{tabular}{ll}
\toprule
Category & Constraints \\
\midrule
Precedence &
  \texttt{start\_at\_start}, \texttt{start\_at\_end}, \texttt{end\_at\_start}, \texttt{end\_at\_end}, \\
  & \texttt{start\_before\_start}, \texttt{start\_before\_end}, \texttt{end\_before\_start}, \texttt{end\_before\_end} \\
Grouping &
  \texttt{span}, \texttt{alternative}, \texttt{synchronize} \\
Sequence &
  \texttt{SeqNoOverlap}, \texttt{first}, \texttt{last}, \texttt{before}, \texttt{previous}, \\
  & \texttt{same\_sequence}, \texttt{same\_common\_subsequence} \\
Cumulative &
  \texttt{pulse}, \texttt{step\_at}, \texttt{step\_at\_start}, \texttt{step\_at\_end}, \\
  & \texttt{cumul\_range}, \texttt{always\_in}, \texttt{height\_at\_start}, \texttt{height\_at\_end} \\
State function &
  \texttt{always\_equal}, \texttt{always\_in}, \texttt{always\_constant}, \texttt{always\_no\_state}, \\
  & \texttt{requires\_state}, \texttt{sets\_state} \\
Forbidden time &
  \texttt{forbid\_start}, \texttt{forbid\_end}, \texttt{forbid\_extent} \\
Presence &
  \texttt{presence\_implies}, \texttt{presence\_or}, \texttt{presence\_xor}, \\
  & \texttt{all\_present\_or\_all\_absent}, \texttt{presence\_or\_all}, \texttt{if\_present\_then}, \\
  & \texttt{at\_least\_k\_present}, \texttt{at\_most\_k\_present}, \texttt{exactly\_k\_present} \\
Overlap &
  \texttt{must\_overlap}, \texttt{overlap\_at\_least}, \texttt{no\_overlap\_pairwise}, \texttt{disjunctive} \\
Chain &
  \texttt{chain}, \texttt{strict\_chain} \\
Bounds &
  \texttt{release\_date}, \texttt{deadline}, \texttt{time\_window} \\
\bottomrule
\end{tabular}
\end{table}

\begin{table}[t]
\centering
\caption{Expressions provided by \pycspsched{}.}
\label{tab:expressions}
\footnotesize
\setlength{\tabcolsep}{4pt}
\begin{tabular}{ll}
\toprule
Category & Expressions \\
\midrule
Interval accessors &
  \texttt{start\_of}, \texttt{end\_of}, \texttt{size\_of}, \texttt{length\_of}, \texttt{presence\_of}, \\
  & \texttt{overlap\_length}, \texttt{expr\_min}, \texttt{expr\_max} \\
Sequence (next) &
  \texttt{next\_arg}, \texttt{start\_of\_next}, \texttt{end\_of\_next}, \texttt{size\_of\_next}, \texttt{length\_of\_next} \\
Sequence (prev) &
  \texttt{prev\_arg}, \texttt{start\_of\_prev}, \texttt{end\_of\_prev}, \texttt{size\_of\_prev}, \texttt{length\_of\_prev} \\
Aggregate &
  \texttt{count\_present}, \texttt{earliest\_start}, \texttt{latest\_end}, \texttt{span\_length}, \texttt{makespan} \\
Element indexing &
  \texttt{ElementArray}, \texttt{ElementMatrix}, \texttt{element}, \texttt{element2d} \\
\bottomrule
\end{tabular}
\end{table}

\paragraph{Element expressions.}
Scheduling objectives frequently depend on data looked up through a variable index: for instance, the setup cost between 2 consecutive tasks depends on which task comes next, an identity known only at solve time.
\texttt{ElementArray} and \texttt{ElementMatrix} provide this capability: \texttt{M[i, next\_arg(seq, x)]} returns a \pycsp{} element expression that indexes into a cost matrix using the (variable) successor identity of interval~\texttt{x} in sequence~\texttt{seq}.
The library compiles these into \xcsp{} \texttt{element} constraints.
This pattern is central to models with sequence-dependent setup costs (e.g., LotSizing) and routing objectives.

\subsection{Mixing scheduling abstractions with raw \pycsp{}}
\label{sec:escape-hatch}

The scheduling layer is not a closed sublanguage: \pycspsched{} is embedded in \pycsp{}, so high-level scheduling constructs and ordinary \pycsp{} variables/constraints can be mixed in the same model and posted in the same \texttt{satisfy()} block.
The accessors \texttt{start\_of(x)}, \texttt{end\_of(x)}, \texttt{size\_of(x)}, \texttt{length\_of(x)}, and \texttt{presence\_of(x)} return \pycsp{} expressions over the underlying integer/Boolean variables, so any constraint expressible in \pycsp{} can be posted on intervals --- even when no scheduling constraint covers the desired pattern directly.

For example, suppose a machine must be \emph{cleaned after every 5 executed tasks} on a sequence \texttt{seq=[t1,...,tn]} of optional intervals.
This is not a built-in scheduling constraint, but it can be modeled by combining \texttt{SeqNoOverlap} for the temporal/resource structure with a counting constraint over presence variables and a chain of cleaning intervals \texttt{c1,...,ck}:

\noindent\begin{minipage}{\linewidth}
\begin{lstlisting}[style=pycode]
exec = [presence_of(t) for t in seq]                # Boolean presences
satisfy(
    SeqNoOverlap(SequenceVar(seq + cleaning)),      # machine occupancy
    [Sum(exec[:i+1]) - 5*j == 0                     # cleaning fires every 5 tasks
        for j, c in enumerate(cleaning, start=1)
        for i in range(len(seq))
        if start_of(c) >= end_of(seq[i])],
    [size_of(c) == clean_duration for c in cleaning]
)
\end{lstlisting}
\end{minipage}

The scheduling abstraction handles the regular machine-occupancy structure, while raw \pycsp{} expresses the side counting rule.
This pattern --- compose, do not subclass --- is what we mean by saying the abstraction can be \emph{broken} when needed: model authors fall back to ordinary \pycsp{} only at the points where a problem-specific constraint exceeds the library's vocabulary, and the rest of the model still benefits from scheduling-aware compilation.

\paragraph{Visualization.}
The library includes a \texttt{visu} module for rendering scheduling solutions as Gantt charts.
A \texttt{timeline} is composed of named \texttt{panel}s, each displaying interval rectangles, transition markers, pause periods, or cumulative-function profiles.
Solved intervals can be plotted directly via \texttt{show\_interval} and \texttt{show\_sequence}, and the output can be displayed interactively or saved to file.
This module is intended for debugging and result presentation rather than solver interaction.

\paragraph{Compilation to \pycsp{}/\xcsp{}.}
\xcsp{} solvers (ACE, Choco, CoSoCo) operate exclusively on the classical \xcsp{} fragment and have no notion of interval, sequence, presence, or intensity; the scheduling abstractions therefore live entirely at the modeling layer of \pycspsched{} and never reach the solver, ensuring full compatibility with the existing \pycsp{}/\xcsp{} solver ecosystem.
Each scheduling object is lowered to standard \pycsp{} variables and constraints, reusing existing \pycsp{} globals (\texttt{NoOverlap}, \texttt{Cumulative}) when the scheduling pattern matches their signature.
An \texttt{IntervalVar} becomes integer variables for start, end, length, and size, plus a Boolean presence variable when the interval is optional.

Temporal constraints on optional intervals are guarded by presence literals.
For instance, any temporal relation \(\phi\) involving optional interval \(x_i\) is compiled as:
\[
(\texttt{presence}_i = 0) \;\lor\; \phi.
\]
This preserves satisfiability when an optional activity is absent.

The compilation of \texttt{SeqNoOverlap} depends on the features used and offers two complementary fallback strategies when the \xcsp{} \texttt{noOverlap} global cannot be emitted directly.

\textit{(i) Direct \texttt{noOverlap}.}
When all intervals are mandatory and no transition matrix is specified, the library emits the \xcsp{} \texttt{noOverlap} global constraint, allowing the solver to apply dedicated propagation (e.g., theta-tree, edge-finding).

\textit{(ii) Pairwise disjunctive decomposition.}
When intervals are optional or transition times are present, \xcsp{} \texttt{noOverlap} does not support presence guards or type-based transitions, so the library falls back by default to \(O(n^2)\) pairwise disjunctions:
\[
(\neg p_i) \lor (\neg p_j) \lor (e_i \le s_j) \lor (e_j \le s_i), \quad \forall\; i < j,
\]
where \(s_i, e_i, p_i\) denote the start, end, and presence of interval~\(x_i\).
When transition times \(d_{ij}\) are specified, the temporal disjuncts become \(e_i + d_{ij} \le s_j\) and \(e_j + d_{ji} \le s_i\).
This decomposition denies the solver access to efficient global propagators and is the likely cause of the performance regression observed in MRCPSP (Section~\ref{sec:discussion}).

\textit{(iii) Time-based unary cumulative.}
For optional intervals without transitions, an alternative compilation path mimics \texttt{noOverlapOptional} via \texttt{Cumulative} on a unit-capacity resource: each interval contributes its presence variable \(p_i \in \{0,1\}\) as height on a resource of capacity~1.
Formally, \texttt{Cumulative}\((s_i, \text{size}_i, p_i)_{i \in [n]} \le 1\) enforces non-overlap because at most one present interval can occupy any time point.
\xcsp{} \texttt{Cumulative} accepts variable heights, so this is a valid target; it gives the solver access to time-tabling and sweep propagation rather than \(O(n^2)\) Boolean reasoning.
We discuss this alternative further in Section~\ref{sec:discussion}; it is a planned compilation option.

\textit{Cumulative profiles.}
The \texttt{Cumulative} global is also emitted directly when the cumulative function is a simple sum of fixed-height pulses with an upper-bound capacity.
More complex profiles (variable heights, step functions, range constraints) are decomposed into primitive constraints.

\textit{MRCPSP regression in detail.}
MRCPSP combines two patterns that interact poorly with the current compilation: (a) optional alternative-mode intervals over a single underlying activity, and (b) machine-disjunctive constraints aggregating those modes.
The hand-crafted classical model exploits problem structure by introducing one start variable per activity and one mode-selection variable per activity, posting a global cumulative directly over those.
The transformed-from-scheduling model instead introduces one optional interval per (activity, mode) pair and posts \texttt{SeqNoOverlap} over them, which currently falls back to pairwise disjunctions.
The variable count is similar (+40\% augmentation), but the constraint structure replaces a single global with \(O(n^2)\) presence-guarded disjuncts, which is what makes ACE slower on this family.
Strategy~(iii) above, or an \xcsp{} extension of \texttt{noOverlap} carrying presence and transition guards, is the principled fix.

Intensity profiles are handled by precomputing the set of feasible tuples \((\textit{start}, \textit{size}, \textit{length})\) that satisfy the discretized integral equation.
For the profile in Figure~\ref{fig:interval-docstyle} (100\% intensity before \(t{=}10\), then 50\%), a task with \(\textit{size}{=}10\) yields tuples such as \((0,10,10)\), \((5,10,15)\), and \((10,10,20)\): as the task starts later relative to the high-intensity region, its elapsed length grows.
These tuples are posted as an extensional constraint, enforcing start-length consistency without runtime scanning.


\section{Experimental Protocol}
\label{sec:protocol}

The experiment addresses 3 questions:
\begin{enumerate}
\item Does the scheduling formulation preserve solution quality?
\item What is the runtime cost or benefit of compilation-generated constraints?
\item How does structural augmentation (variable/constraint count) vary across problem families?
\end{enumerate}

We evaluate 261 paired instances drawn from 17 model families:
AircraftLanding~(13), BACP~(28), BusScheduling~(12), CyclicRCPSP~(10), EchelonStock2~(10), FlexibleJobshop~(5), FlexibleJobshopScen~(9), Flow-shop~(2), Job-shop~(2), LotSizing~(59), MRCPSP~(10), MSPSP~(6), NursingWorkload~(12), RCPSP~(30), Rehearsal~(2), TestScheduling~(31), and TravelingTournamentWithPredefinedVenues~(20).
A description of each problem can be found in CSPLib~\cite{csplib}\footnote{\url{https://www.csplib.org/Problems/categories.html}}.
Instance sources include PSPLIB~\cite{kolisch1997psplib,psplib}, CSPLib~\cite{csplib,csplib61}, \xcsp{} challenge datasets~\cite{xcsp3}, and MiniZinc challenge datasets~\cite{mznchallenge2023}.
Table~\ref{tab:model-features} details the scheduling abstractions used by each model.

\begin{table}[t]
\centering
\caption{Scheduling abstractions used by each model family. IV = \texttt{IntervalVar}, SV = \texttt{SequenceVar}, Opt = at least one interval declared with \texttt{optional=True} (presence guards in compilation), ebs = \texttt{end\_before\_start}, alt = \texttt{alternative}, SNO = \texttt{SeqNoOverlap}, SC = \texttt{SeqCumulative}, st = \texttt{start\_time}, et = \texttt{end\_time}, pt = \texttt{presence\_time}, cp = \texttt{count\_present}, na = \texttt{next\_arg}, EM = \texttt{ElementMatrix}.}
\label{tab:model-features}
\scriptsize
\setlength{\tabcolsep}{2.5pt}
\begin{tabular}{@{}l ccc cccc cccccc@{}}
\toprule
 & \multicolumn{3}{c}{Variables} & \multicolumn{4}{c}{Constraints} & \multicolumn{6}{c}{Expressions} \\
\cmidrule(lr){2-4} \cmidrule(lr){5-8} \cmidrule(lr){9-14}
Family & IV & SV & Opt & ebs & alt & SNO & SC & st & et & pt & cp & na & EM \\
\midrule
AircraftLanding       & $\bullet$ &           &           &           &           &           &           & $\bullet$ &           &           &           &           &           \\
BACP                  & $\bullet$ &           &           & $\bullet$ &           &           &           & $\bullet$ &           &           &           &           &           \\
BusScheduling         & $\bullet$ &           & $\bullet$ &           &           &           &           &           &           & $\bullet$ & $\bullet$ &           &           \\
CyclicRCPSP           & $\bullet$ &           &           &           &           &           & $\bullet$ & $\bullet$ & $\bullet$ &           &           &           &           \\
EchelonStock2         & $\bullet$ & $\bullet$ & $\bullet$ &           &           & $\bullet$ &           &           &           & $\bullet$ &           &           &           \\
FlexibleJobshop       & $\bullet$ &           & $\bullet$ & $\bullet$ & $\bullet$ &           & $\bullet$ &           & $\bullet$ &           &           &           &           \\
FlexibleJobshopScen   & $\bullet$ &           & $\bullet$ & $\bullet$ & $\bullet$ &           & $\bullet$ &           & $\bullet$ & $\bullet$ &           &           &           \\
Flow-shop             & $\bullet$ & $\bullet$ &           &           &           & $\bullet$ &           & $\bullet$ & $\bullet$ &           &           &           &           \\
Job-shop              & $\bullet$ & $\bullet$ &           &           &           & $\bullet$ &           & $\bullet$ & $\bullet$ &           &           &           &           \\
LotSizing             & $\bullet$ & $\bullet$ &           & $\bullet$ &           & $\bullet$ &           & $\bullet$ &           &           &           & $\bullet$ & $\bullet$ \\
MRCPSP                & $\bullet$ &           & $\bullet$ & $\bullet$ & $\bullet$ &           &           & $\bullet$ & $\bullet$ & $\bullet$ &           &           &           \\
MSPSP                 & $\bullet$ &           &           & $\bullet$ &           &           & $\bullet$ & $\bullet$ & $\bullet$ &           &           &           &           \\
NursingWorkload       & $\bullet$ &           &           &           &           &           &           &           & $\bullet$ &           &           &           &           \\
RCPSP                 & $\bullet$ &           &           & $\bullet$ &           &           & $\bullet$ & $\bullet$ &           &           &           &           &           \\
Rehearsal             & $\bullet$ & $\bullet$ &           &           &           & $\bullet$ &           & $\bullet$ & $\bullet$ &           &           &           &           \\
TestScheduling        & $\bullet$ & $\bullet$ &           &           &           & $\bullet$ & $\bullet$ & $\bullet$ & $\bullet$ &           &           &           &           \\
TTPV                  & $\bullet$ &           &           &           &           &           &           &           & $\bullet$ &           &           &           &           \\
\bottomrule
\end{tabular}
\end{table}

Each instance is solved twice, once with the classical \pycsp{} formulation (from \pycsp{}-models\footnote{\url{https://github.com/xcsp3team/pycsp3-models/}}) and once with the scheduling formulation, using the same solver and timeout.
The solver is ACE~2.5~\cite{ace_solver} (bundled \texttt{ACE-2.5.jar}), invoked through \pycsp{} with a 1200\,s timeout per run on a single process (no parallelism).
The runtime environment is Python~3.12.6, OpenJDK~24 on an Intel Core i7-13700 (16~cores, 80\,GB RAM, Windows~11).

Each pair is run 5 times; results are averaged over repetitions.
We report descriptive paired statistics (status counts, speedup distributions, objective comparisons) and treat objective differences outside doubly-proved optimal pairs as best-found-solution snapshots rather than correctness indicators.

\paragraph{Solver status.}
Each run terminates with one of four \emph{solver statuses} reported by ACE within the 1200\,s budget:
\textsc{Optimum} (the solver proves optimality and returns the optimal objective),
\textsc{Sat} (the solver finds at least one feasible solution but does not prove optimality before timeout),
\textsc{Unsat} (the solver proves the instance has no feasible solution), and
\textsc{Timeout} (the solver returns no information beyond the budget being exhausted).
We say a pair has \emph{status agreement} when the classical and scheduling formulations report the same status; comparable objectives are restricted to pairs where both runs return at least one feasible solution.

\section{Results}
\label{sec:results}

All results below are averaged over 5 independent runs per pair.

\paragraph{Aggregate outcomes.}
Table~\ref{tab:global} summarizes the 261 instance pairs.
The 2 formulations agree on solver status in 80.8\% of pairs (211/261).
Among the 72~pairs where both formulations prove optimality, all objective values match exactly, confirming that the compilation introduces no semantic divergence on fully solved instances.
The scheduling formulation is faster on 62 of 261 pairs (23.8\%), the classical formulation is faster on 78 (29.9\%), and 121 (46.4\%) are runtime ties at the recorded precision; the mean time saving is 137.58\,s though the median speedup is 1.00\(\times\).

\begin{table}[t]
\centering
\caption{Global comparison over 261 classical/scheduling pairs.}
\label{tab:global}
\begin{tabular}{lr}
\toprule
Metric & Value \\
\midrule
Status agreement & 211/261 (80.8\%) \\
Both successful statuses (SAT/OPT/UNSAT) & 199/261 \\
Comparable objectives & 196 pairs \\
Objective: classical better / scheduling better / tie & 28 / 23 / 145 \\
Both OPTIMUM with objective available & 72 pairs \\
Exact objective match (both-OPTIMUM) & 72/72 (100\%) \\
Pairs where scheduling solve time is lower & 62/261 (23.8\%) \\
Median solve speedup (classical/scheduling) & 1.00\(\times\) \\
Mean solve-time difference (scheduling $-$ classical) & $-$137.58 s \\
\bottomrule
\end{tabular}
\end{table}

\paragraph{Per-family analysis.}
Most families cluster near the diagonal in the family-level solve-time scatter (Figure~\ref{fig:solve-scatter} in Appendix~\ref{app:scatter}); the per-family numbers in Table~\ref{tab:structural-outliers} read off this scatter.
3 families stand out as clear scheduling wins: AircraftLanding (3.47\(\times\) speedup), Rehearsal (3.31\(\times\)), and MSPSP (5.82\(\times\)).
3 families regress: LotSizing (0.29\(\times\)), MRCPSP (0.61\(\times\)), and FlexibleJobshopScen (0.89\(\times\)).

8 of 17 families have zero or negligible structural augmentation ($<$1\% change in both variables and constraints).
Table~\ref{tab:structural-outliers} reports the remaining 9 families, including 2 where the scheduling formulation is \emph{smaller} (BACP and Rehearsal).

\begin{table}[t]
\centering
\caption{Families with non-zero structural augmentation. TTPV abbreviates TravelingTournamentWithPredefinedVenues. Negative augmentation means the scheduling formulation is smaller. \emph{Speedup} is the family-mean ratio classical/scheduling solve time. \emph{Obj Better} indicates which formulation reaches the better objective on the family (\textit{Classical} / \textit{Scheduling} / \textit{Tie}).}
\label{tab:structural-outliers}
\begin{tabular}{lrrrrr}
\toprule
Family & Inst. & Var Aug. (\%) & Ctr Aug. (\%) & Speedup & Obj Better \\
\midrule
LotSizing & 59 & $-$6.9 & +109.8 & 0.29 & Classical \\
FlexibleJobshopScen & 9 & +268.7 & +383.7 & 0.89 & Classical \\
NursingWorkload & 12 & +320.7 & +17.7 & 0.95 & Classical \\
EchelonStock2 & 10 & +100.0 & +219.7 & 1.00 & Tie \\
FlexibleJobshop & 5 & +47.5 & +42.5 & 0.97 & Scheduling \\
TTPV & 20 & +43.3 & +11.8 & 1.00 & Classical \\
MRCPSP & 10 & +40.3 & +30.4 & 0.61 & Classical \\
BACP & 28 & $-$87.7 & $-$37.4 & 1.16 & Tie \\
Rehearsal & 2 & $-$69.2 & $-$89.3 & 3.31 & Tie \\
\bottomrule
\end{tabular}
\end{table}

\paragraph{Status disagreements.}
Of the 50 disagreeing pairs (19.2\%), the breakdown by status combination is shown in Table~\ref{tab:status-disagreements}.
Timeout-related cases account for 34 of 50 (68.0\%); LotSizing alone contributes 38 of the 50~disagreements.
These patterns indicate search-progress differences under timeout, not semantic inconsistencies.

\begin{table}[t]
\centering
\caption{Status disagreement breakdown for the 50 pairs (19.2\%) where the two formulations report different solver statuses.}
\label{tab:status-disagreements}
\begin{tabular}{llr}
\toprule
Classical status & Scheduling status & Pairs \\
\midrule
\textsc{Sat}     & \textsc{Timeout} & 32 \\
\textsc{Optimum} & \textsc{Sat}     & 11 \\
\textsc{Sat}     & \textsc{Optimum} &  5 \\
\textsc{Timeout} & \textsc{Sat}     &  2 \\
\midrule
\multicolumn{2}{l}{Total} & 50 \\
\bottomrule
\end{tabular}
\end{table}

\paragraph{Summary statistics.}
At family granularity (equal weight over 17~families), the average augmentation is +38.6\% variables and +40.5\% constraints, with a mean speedup of 1.50\(\times\).
At pair granularity (all 261~pairs), the median speedup is 1.00\(\times\) (for positive-speedup pairs, \(q_{10}{=}0.97\), \(q_{90}{=}1.22\)).
The formulation layer is performance-neutral on most instances, with concentrated gains and losses in specific families.

\section{Discussion}
\label{sec:discussion}

The results reveal a clear split between families where the scheduling formulation helps and families where it hurts, and the reasons are traceable to formulation structure.

\paragraph{Why some families improve.}
MSPSP achieves a 5.82\(\times\) speedup with zero augmentation: the scheduling and classical models compile to the same \xcsp{} constraints, but the scheduling model is generated from a more structured representation that produces a more solver-friendly variable ordering.
We treat this number with caution: because the underlying constraint set is byte-for-byte the same, the gain is not attributable to the scheduling abstractions themselves but to a side-effect of how \pycspsched{} declares variables (interval-by-interval, interleaving start/end/length) versus the classical model (all start times first, then all end times).
Under ACE's default variable-selection heuristic, the interleaved order yields better branching choices early in search, which is properly an artifact of the modeler's variable-creation pattern rather than of the scheduling layer.
A solver-side neutralization (e.g., a fixed lex variable order or a more autonomous heuristic) is the principled way to remove this confound; we plan such an ablation as future work and refrain from claiming this speedup as a genuine abstraction win.
AircraftLanding and Rehearsal show a different pattern: the scheduling formulation is \emph{both smaller and faster}, because interval abstractions capture the problem's temporal structure without the redundant channeling variables introduced by the classical encoding --- a gain that is not attributable to variable ordering.

\paragraph{Why some families regress.}
LotSizing now exhibits a milder structural profile ($-$6.9\% variables, +109.8\% constraints), but still regresses strongly in runtime (0.29\(\times\) speedup) and contributes most disagreements (38 of 50).
This suggests that the dominant issue is no longer variable explosion but the constraint-level encoding and propagation strength under timeout on the hardest instances.
FlexibleJobshopScen inflates both variables (+268.7\%) and constraints (+383.7\%) by duplicating optional interval decisions across scenarios, a structure the classical model avoids.
MRCPSP regresses to 0.61\(\times\) speedup despite moderate augmentation (+40.3\% variables).
Because MRCPSP uses optional intervals (tasks with alternative execution modes), the \texttt{SeqNoOverlap} constraint falls back to \(O(n^2)\) pairwise disjunctions instead of emitting the \xcsp{} \texttt{noOverlap} global, preventing the solver from using efficient sweeping or theta-tree propagation.
Extending the compilation to emit \texttt{noOverlap} with optionality support, or proposing an \xcsp{} extension for it, is a clear avenue for reducing this regression.


\section{Conclusion}
\label{sec:conclusion}

\pycspsched{} adds interval and sequence abstractions to \pycsp{} while preserving the complete separation between modeling and solving that underpins the \pycsp{}/\xcsp{} ecosystem~\cite{pycsp3_2020}.
The compilation reuses existing \pycsp{} globals (\texttt{NoOverlap}, \texttt{Cumulative}) when applicable and falls back to intension constraints otherwise.
Correctness is verified by exact objective agreement on all 72 doubly-optimal pairs, and runtime impact is mixed across families, with concentrated regressions traceable to specific formulation choices (LotSizing, FlexibleJobshopScen, MRCPSP).

The most promising directions for future work are emitting \xcsp{} global constraints (e.g., \texttt{noOverlap} with optionality) instead of pairwise decompositions, reducing the structural overhead in outlier families through tighter compilation rules, and extending the benchmark corpus with additional model families.

The library is released under the MIT licence and will be available on PyPI.
The benchmark pipeline, model code, and all reported artifacts are available at \url{https://github.com/sohaibafifi/pycsp3-scheduling}.

\bibliography{references}

\appendix

\section{Solve-time Scatter}
\label{app:scatter}

\begin{figure}[!t]
\centering
\includegraphics[width=0.76\textwidth]{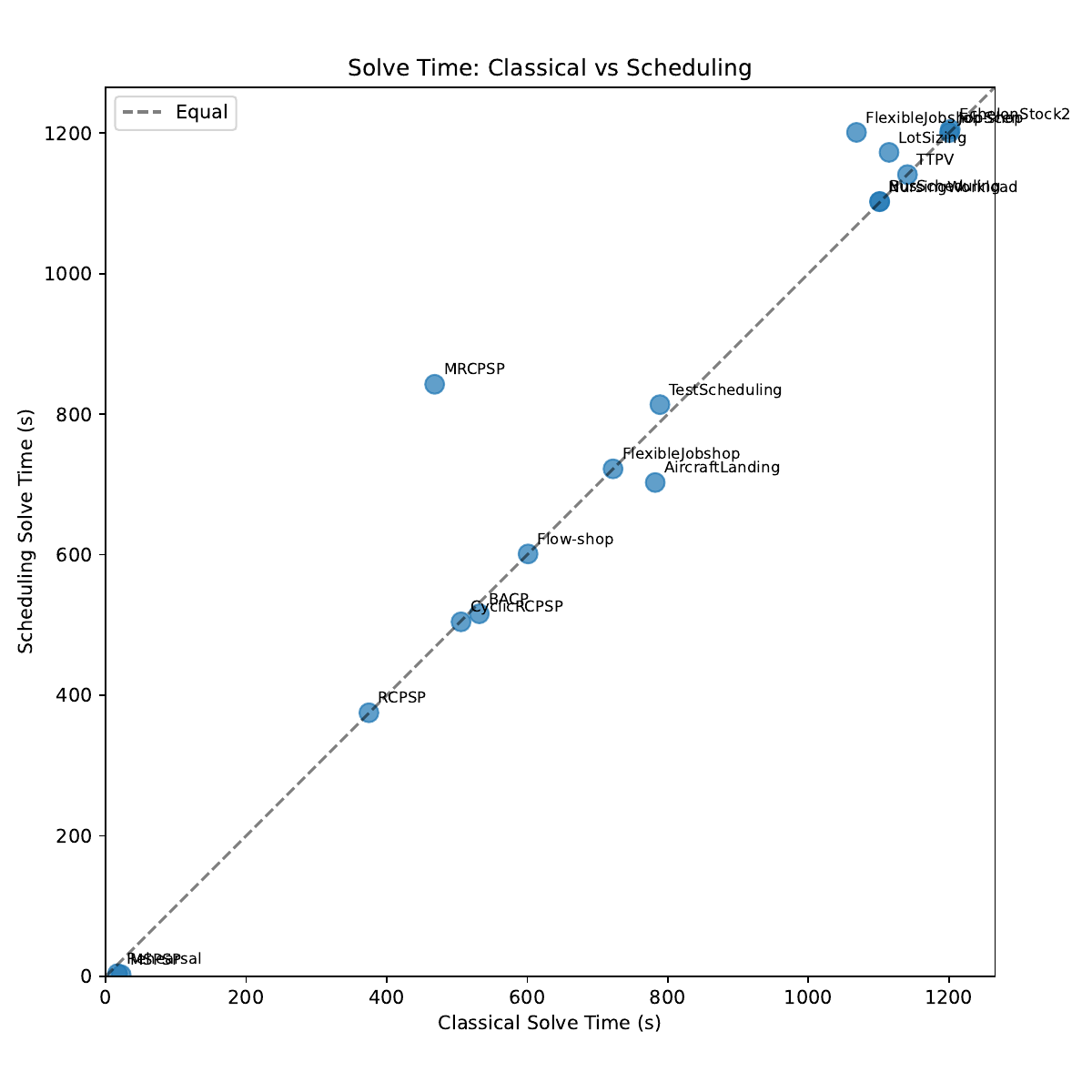}
\caption{Family-level solve-time scatter (classical vs.\ scheduling). Points below the diagonal indicate faster scheduling runs. Numerical summaries are reported per family in Table~\ref{tab:structural-outliers}.}
\label{fig:solve-scatter}
\end{figure}

\section{Constraint and Expression Reference}
\label{app:reference}

This appendix provides a complete description of all constraints and expressions available in \pycspsched{}.
In all descriptions, \(a\) and \(b\) denote interval variables; \(s\), \(e\), \(p\) denote start, end, and presence respectively.

\subsection{Precedence Constraints}

\begin{description}
\item[\texttt{start\_at\_start(a, b, delay)}] Enforces \(s(b) = s(a) + \textit{delay}\).
\item[\texttt{start\_at\_end(a, b, delay)}] Enforces \(s(b) = e(a) + \textit{delay}\).
\item[\texttt{end\_at\_start(a, b, delay)}] Enforces \(e(a) = s(b) + \textit{delay}\).
\item[\texttt{end\_at\_end(a, b, delay)}] Enforces \(e(b) = e(a) + \textit{delay}\).
\item[\texttt{start\_before\_start(a, b, delay)}] Enforces \(s(b) \ge s(a) + \textit{delay}\).
\item[\texttt{start\_before\_end(a, b, delay)}] Enforces \(e(b) \ge s(a) + \textit{delay}\).
\item[\texttt{end\_before\_start(a, b, delay)}] Enforces \(s(b) \ge e(a) + \textit{delay}\). This is the classic precedence constraint.
\item[\texttt{end\_before\_end(a, b, delay)}] Enforces \(e(b) \ge e(a) + \textit{delay}\).
\end{description}

All precedence constraints default to \(\textit{delay}=0\) and are guarded by presence when the intervals are optional.

\subsection{Grouping Constraints}

\begin{description}
\item[\texttt{span(main, subtasks)}] Constrains \(s(\textit{main}) = \min_i s(i)\) and \(e(\textit{main}) = \max_i e(i)\) over all present subtasks.
\item[\texttt{alternative(main, alternatives, cardinality)}] Selects exactly \textit{cardinality} alternatives whose start and end times match \textit{main}. Default cardinality is~1.
\item[\texttt{synchronize(main, intervals)}] All present intervals must have the same start and end times as \textit{main}.
\end{description}

\subsection{Sequence Constraints}

\begin{description}
\item[\texttt{SeqNoOverlap(seq, transition\_matrix, is\_direct)}] No two present intervals in the sequence overlap. When a transition matrix \(d_{ij}\) is provided, a minimum gap of \(d_{ij}\) is enforced between consecutive intervals of types \(i\) and \(j\). With \texttt{is\_direct=True}, transitions apply only to immediate successors.
\item[\texttt{first(seq, interval)}] The interval must start before all other present intervals in the sequence.
\item[\texttt{last(seq, interval)}] The interval must end after all other present intervals in the sequence.
\item[\texttt{before(seq, a, b)}] Enforces \(e(a) \le s(b)\) within the sequence ordering.
\item[\texttt{previous(seq, a, b)}] Enforces that \(a\) immediately precedes \(b\) with no other present interval between them.
\item[\texttt{same\_sequence(seq1, seq2)}] Common intervals maintain the same position in both sequences.
\item[\texttt{same\_common\_subsequence(seq1, seq2)}] Common intervals maintain the same relative order in both sequences (weaker than \texttt{same\_sequence}).
\end{description}

\subsection{Cumulative Functions}

\begin{description}
\item[\texttt{pulse(interval, height)}] Rectangular resource usage of \textit{height} during the interval's execution.
\item[\texttt{step\_at(time, height)}] Permanent step change of \textit{height} at a fixed time point.
\item[\texttt{step\_at\_start(interval, height)}] Permanent step change of \textit{height} at \(s(\textit{interval})\).
\item[\texttt{step\_at\_end(interval, height)}] Permanent step change of \textit{height} at \(e(\textit{interval})\).
\item[\texttt{cumul\_range(cumul, min, max)}] Enforces \(\textit{min} \le \textit{cumul}(t) \le \textit{max}\) at all time points.
\item[\texttt{always\_in(cumul, range, min, max)}] Enforces \(\textit{min} \le \textit{cumul}(t) \le \textit{max}\) during a specified interval or fixed time range.
\item[\texttt{height\_at\_start(interval, cumul)}] Expression for the cumulative value at \(s(\textit{interval})\).
\item[\texttt{height\_at\_end(interval, cumul)}] Expression for the cumulative value at \(e(\textit{interval})\).
\end{description}

Cumulative expressions are combined with \texttt{+} and constrained with \texttt{<=} to model capacity limits.

\subsection{State Function Constraints}

\begin{description}
\item[\texttt{always\_equal(func, interval, value)}] The state function equals \textit{value} throughout the interval.
\item[\texttt{always\_in(func, interval, min, max)}] The state function stays within \([\textit{min}, \textit{max}]\) during the interval.
\item[\texttt{always\_constant(func, interval)}] The state function does not change value during the interval.
\item[\texttt{always\_no\_state(func, interval)}] The state function has no defined state during the interval.
\item[\texttt{requires\_state(interval, func, state)}] The interval requires the state function to equal \textit{state} (delegates to \texttt{always\_equal}).
\item[\texttt{sets\_state(interval, func, before, after)}] Models a state transition: the function equals \textit{before} at the start and \textit{after} at the end.
\end{description}

\subsection{Forbidden Time Constraints}

\begin{description}
\item[\texttt{forbid\_start(interval, periods)}] The interval cannot start during any forbidden period \([s, e)\).
\item[\texttt{forbid\_end(interval, periods)}] The interval cannot end during any forbidden period \((s, e]\).
\item[\texttt{forbid\_extent(interval, periods)}] The interval cannot overlap any forbidden period.
\end{description}

\subsection{Presence Constraints}

\begin{description}
\item[\texttt{presence\_implies(a, b)}] Enforces \(p(a) \Rightarrow p(b)\).
\item[\texttt{presence\_or(a, b)}] Enforces \(p(a) \lor p(b)\).
\item[\texttt{presence\_xor(a, b)}] Enforces \(p(a) \oplus p(b)\) (exactly one is present).
\item[\texttt{all\_present\_or\_all\_absent(intervals)}] All intervals share the same presence status.
\item[\texttt{presence\_or\_all(*intervals)}] At least one interval is present.
\item[\texttt{if\_present\_then(interval, ctr)}] The constraint \textit{ctr} is active only when the interval is present.
\item[\texttt{at\_least\_k\_present(intervals, k)}] Enforces \(\sum p(i) \ge k\).
\item[\texttt{at\_most\_k\_present(intervals, k)}] Enforces \(\sum p(i) \le k\).
\item[\texttt{exactly\_k\_present(intervals, k)}] Enforces \(\sum p(i) = k\).
\end{description}

\subsection{Overlap and Disjunctive Constraints}

\begin{description}
\item[\texttt{must\_overlap(a, b)}] Enforces \(s(a) < e(b) \land s(b) < e(a)\).
\item[\texttt{overlap\_at\_least(a, b, min)}] Enforces \(\min(e(a), e(b)) - \max(s(a), s(b)) \ge \textit{min}\).
\item[\texttt{no\_overlap\_pairwise(intervals)}] Pairwise non-overlap: for each pair \((i,j)\), \(e(i) \le s(j)\) or \(e(j) \le s(i)\).
\item[\texttt{disjunctive(intervals, transition\_times)}] Unary resource: at most one interval active at any time, with optional transition times.
\end{description}

\subsection{Chain and Bounds Constraints}

\begin{description}
\item[\texttt{chain(intervals, delays)}] Sequential execution: \(e(\textit{intervals}[i]) + \textit{delays}[i] \le s(\textit{intervals}[i{+}1])\).
\item[\texttt{strict\_chain(intervals, delays)}] Strict chain with no gaps: \(e(\textit{intervals}[i]) + \textit{delays}[i] = s(\textit{intervals}[i{+}1])\).
\item[\texttt{release\_date(interval, time)}] Enforces \(s(\textit{interval}) \ge \textit{time}\).
\item[\texttt{deadline(interval, time)}] Enforces \(e(\textit{interval}) \le \textit{time}\).
\item[\texttt{time\_window(interval, earliest, latest)}] Enforces \(s \ge \textit{earliest}\) and \(e \le \textit{latest}\).
\end{description}

\subsection{Interval Expressions}

\begin{description}
\item[\texttt{start\_of(interval, absent\_value)}] Start time of the interval, or \textit{absent\_value} if absent.
\item[\texttt{end\_of(interval, absent\_value)}] End time of the interval, or \textit{absent\_value} if absent.
\item[\texttt{size\_of(interval, absent\_value)}] Duration of the interval, or \textit{absent\_value} if absent.
\item[\texttt{length\_of(interval, absent\_value)}] Length of the interval (may differ from size with intensity functions), or \textit{absent\_value} if absent.
\item[\texttt{presence\_of(interval)}] Boolean expression (0~or~1) indicating whether the interval is present.
\item[\texttt{overlap\_length(a, b)}] Duration of overlap: \(\max(0,\; \min(e(a), e(b)) - \max(s(a), s(b)))\).
\item[\texttt{expr\_min(*args)}] Minimum of multiple interval expressions or constants.
\item[\texttt{expr\_max(*args)}] Maximum of multiple interval expressions or constants.
\end{description}

\subsection{Sequence Expressions}

All sequence expressions take a \textit{last\_value} (returned when the interval is last in the sequence) and an \textit{absent\_value} (returned when the interval is absent).

\begin{description}
\item[\texttt{next\_arg(seq, interval)}] Type/ID of the next interval in the sequence order.
\item[\texttt{start\_of\_next(seq, interval)}] Start time of the next interval.
\item[\texttt{end\_of\_next(seq, interval)}] End time of the next interval.
\item[\texttt{size\_of\_next(seq, interval)}] Size of the next interval.
\item[\texttt{length\_of\_next(seq, interval)}] Length of the next interval.
\item[\texttt{prev\_arg(seq, interval)}] Type/ID of the previous interval (\textit{first\_value} when first).
\item[\texttt{start\_of\_prev(seq, interval)}] Start time of the previous interval.
\item[\texttt{end\_of\_prev(seq, interval)}] End time of the previous interval.
\item[\texttt{size\_of\_prev(seq, interval)}] Size of the previous interval.
\item[\texttt{length\_of\_prev(seq, interval)}] Length of the previous interval.
\end{description}

\subsection{Aggregate and Element Expressions}

\begin{description}
\item[\texttt{count\_present(intervals)}] Number of present intervals: \(\sum p(i)\).
\item[\texttt{earliest\_start(intervals)}] Minimum start time over all present intervals.
\item[\texttt{latest\_end(intervals)}] Maximum end time over all present intervals.
\item[\texttt{span\_length(intervals)}] \(\max(e(i)) - \min(s(i))\) over present intervals.
\item[\texttt{makespan(intervals)}] Latest end time (alias for \texttt{latest\_end}).
\item[\texttt{ElementArray(data)}] 1D array supporting variable-index access: \texttt{array[expr]}.
\item[\texttt{ElementMatrix(matrix)}] 2D matrix supporting \texttt{M[row\_expr, col\_expr]} with boundary handling for last-in-sequence and absent intervals.
\item[\texttt{element(array, index)}] Element expression for \texttt{array[index]} where \textit{index} is a variable.
\item[\texttt{element2d(matrix, row, col)}] Element expression for \texttt{matrix[row][col]} where both indices are variables.
\end{description}

\section{Problem Descriptions}
\label{app:problems}

Brief descriptions of the 17 model families used in the evaluation.
Full problem specifications are available at CSPLib~\cite{csplib}\footnote{\url{https://www.csplib.org/Problems/categories.html}}.

\begin{description}
\item[AircraftLanding] (CSPLib prob011, 13 instances)
Schedule aircraft landings to minimize penalized deviations from target times while respecting separation constraints between consecutive landings.

\item[BACP] (CSPLib prob030, 28 instances)
Balanced Academic Curriculum Problem: assign courses to academic periods while satisfying prerequisites and balancing the course load per period.

\item[BusScheduling] (CSPLib prob022, 12 instances)
Select driver shifts from a predefined set to cover all required bus tasks while minimizing the number of shifts used.

\item[CyclicRCPSP] (MiniZinc Challenge, 10 instances)
Schedule infinitely repeating project tasks with generalized precedence relations and renewable resource constraints to minimize the cycle period.

\item[EchelonStock2] (CSPLib prob040, 10 instances)
Coordinate production and inventory decisions across a multi-level supply chain to minimize holding and setup costs.

\item[FlexibleJobshop] (MiniZinc Challenge, 5 instances)
Assign operations to alternative machines and schedule them to minimize makespan, where each job consists of a sequence of operations with machine flexibility.

\item[FlexibleJobshopScen] (CSPLib prob077, 9 instances)
Stochastic variant: select operation options in a first stage and schedule them under multiple scenarios to minimize expected weighted makespan.

\item[Flow-shop] (\xcsp{} benchmarks, 2 instances)
Schedule jobs through a series of machines where all jobs follow the same machine ordering, minimizing makespan.

\item[Job-shop] (\xcsp{} benchmarks, 2 instances)
Schedule job operations on specific machines where each job has a fixed operation order but different jobs may use machines in different orders, minimizing makespan.

\item[LotSizing] (CSPLib prob058, 59 instances)
Discrete Lot-Sizing and Scheduling: determine production periods for orders on a single machine to minimize the sum of inventory holding costs and sequence-dependent setup costs.

\item[MRCPSP] (MiniZinc Challenge, 10 instances)
Multi-mode Resource-Constrained Project Scheduling: schedule tasks that have alternative execution modes (different durations and resource requirements) under precedence and resource constraints.

\item[MSPSP] (MiniZinc Challenge, 6 instances)
Multi-Skilled Project Scheduling: assign workers with specific skill sets to tasks and schedule them under precedence and skill-coverage constraints.

\item[NursingWorkload] (CSPLib prob069, 12 instances)
Assign patients to nurses while balancing workload distribution across nurses and geographic zones within a hospital ward.

\item[RCPSP] (CSPLib prob061, PSPLIB, 30 instances)
Resource-Constrained Project Scheduling: schedule project tasks under precedence constraints and renewable resource capacities to minimize project duration.

\item[Rehearsal] (CSPLib prob039, 2 instances)
Schedule concert rehearsal pieces to minimize the total waiting time of musicians who are only required for certain pieces.

\item[TestScheduling] (CSPLib prob073, 31 instances)
Schedule industrial tests on machines with resource constraints to minimize the total testing time (makespan).

\item[TravelingTournamentWithPredefinedVenues] (CSPLib prob068, 20 instances)
Schedule tournament rounds with predefined home/away venues to minimize total team travel distance while satisfying round-robin and consecutive-game constraints.
\end{description}

\end{document}